\documentclass[conference,letterpaper, 10pt]{IEEEtran}
\usepackage[top=54pt, left=54pt, right=54pt, bottom=54pt]{geometry}
\usepackage[T1]{fontenc}
\usepackage{cite}
\usepackage{amsmath,amssymb,amsfonts}
\usepackage{graphicx,epstopdf}
\usepackage{graphics}
\usepackage{subcaption}
\usepackage{textcomp}
\usepackage{xcolor}
\usepackage{blindtext}
\usepackage{multirow}
\usepackage{float}
\usepackage{url}           
\usepackage{hyperref}
\usepackage[normalem]{ulem}
\usepackage{afterpage}

\graphicspath{ {./figures/} }
\IEEEoverridecommandlockouts                              

\DeclareUrlCommand{\ULurl}{%
	\renewcommand\UrlLeft{\uline\bgroup}%
	\renewcommand\UrlRight{\egroup}}

\def\BibTeX{{\rm B\kern-.05em{\sc i\kern-.025em b}\kern-.08em
		T\kern-.1667em\lower.7ex\hbox{E}\kern-.125emX}}

\begin{document}
	
	\title{\vspace*{18pt}Model Predictive Control For Multiple Castaway Tracking with an Autonomous Aerial Agent}
	\author{Andreas~Anastasiou\IEEEauthorrefmark{4}\IEEEauthorrefmark{1}, Savvas~Papaioannou\IEEEauthorrefmark{1}, Panayiotis~Kolios\IEEEauthorrefmark{1} and Christos G.~Panayiotou\IEEEauthorrefmark{4}\IEEEauthorrefmark{1}
	\thanks{\IEEEauthorrefmark{1}Authors are with KIOS Research and Innovation Center of Excellence (KIOS CoE),  \IEEEauthorrefmark{4}and with the Department of Electrical and Computer Engineering, University of Cyprus, Nicosia, 1678, Cyprus. E-mail:\texttt{\{anastasiou.antreas, papaioannou.savvas, kolios.panayiotis, christosp\}} @ucy.ac.cy}}
	\maketitle
	
	\begin{abstract}
		Over the past few years, a plethora of advancements in Unmanned Areal Vehicle (UAV) technology has paved the way for UAV-based search and rescue operations with transformative impact to the outcome of critical life-saving missions. This paper dives into the challenging task of multiple castaway tracking using an autonomous UAV agent. Leveraging on the computing power of the modern embedded devices, we propose a Model Predictive Control (MPC) framework for tracking multiple castaways assumed to drift afloat in the aftermath of a maritime accident. We consider a stationary radar sensor that is responsible for signaling the search mission by providing noisy measurements of each castaway's initial state. The UAV agent aims at detecting and tracking the moving targets with its equipped onboard camera sensor that has limited sensing range. In this work, we also experimentally determine the probability of target detection from real-world data by training and evaluating various Convolutional Neural Networks (CNNs). Extensive qualitative and quantitative evaluations demonstrate the performance of the proposed approach.
	\end{abstract}	
	
	\section{Introduction}
	Tracking a target in a sea environment is generally considered a complex task. The mere size of the operation environment results to sparse observations regarding the castaways thus, the estimation error about their positions accumulates. The recent advancements in Unmanned Aerial Vehicle (UAV) technology, has enabled the utilization of UAVs in various  emergency response operations, including search-and-rescue (SAR)~\cite{martinez2021search}\cite{waharte2010supporting}, and search-and-track (SAT)~\cite{wu2020cooperative}\cite{anastasiou2021hyperion} missions. Moreover, UAVs are nowadays being utilized in a plethora of application domains including critical infrastructure inspection \cite{9836051}, surveillance \cite{1677946}, disaster management missions~\cite{terzi2019swifters}\cite{erdelj2017help}, and emergency response~\cite{anastasiou2020swarm}.
	In this paper we envision that the utilization of an autonomous UAV system in maritime scenarios will play a pivotal role in maritime disaster management, security, and safety. It is important to note that during the recent years, the number of lost ships and shipwrecks worldwide, although fortunately decreasing in comparison to a decade ago, is still alarmingly high~\cite{allianz2021lost}. In contrast, the recent problem of migration, which is forcing more people to take the risk of illegal sea travel, has kept the number of castaways precariously high. Those effects are especially evident in the Mediterranean sea. Arrivals through the Mediterranean sea to European countries like Italy, Spain, Greece and Cyprus have exceeded the hundred thousand people and unfortunately, almost two thousand people are currently considered as deceased or missing according to the United Nations High Commissioner for Refugees~\cite{mediterranean2021migrants}.
	
	For the reasons mentioned above, in this work we propose a Model Predictive Control (MPC) framework for tracking multiple castaway targets in maritime environments, using an autonomous UAV agent. The proposed MPC formulation employs a Non-linear Mixed Integer Program (NMIP), for determining the optimal control inputs of an autonomous UAV agent over a finite rolling horizon. This enables accurate multiple castaway tracking and monitoring. Our approach, generates the UAVs control inputs in an on-line fashion so that the overall tracking error, for monitoring the multiple castaway targets, is minimized. Specifically, we consider each castaway's predicted location uncertainty and minimize the distance between the castaway and the UAV's onboard camera sensor, thereby gathering observations with the lowest noise possible. Overall, the key contributions of this work are the following:
	\begin{itemize}
		\item We propose a	Model Predictive Control (MPC) framework, that allows an autonomous UAV agent to track multiple castaways. To achieve this, we formulated a NMIP to compute the optimal control inputs of the UAV over a finite rolling planning horizon, which minimizes the tracking error over multiple targets.
		
		\item We have experimentally analyzed the target detection probability using water buoys, by training various Convolutional Neural Networks (CNNs) from images taken by the UAV from various altitudes. From the evaluation of the CNNs, we constructed a piece-wise linear function that describes the confidence of detection probability based on the UAV's altitude.
		
		\item Finally, we have created a new open-source dataset that consists of aerial images, of water buoys taken with a UAV from various altitudes above the sea level~\cite{antreas_anastasiou_2022_7288444}. Different wave conditions have also been captured in order to create a machine learning object detection algorithm that reflects on the challenges of castaway detection in marine environments.
	\end{itemize}
	
	The rest of this paper is structured as follows. The related work is discussed in Section~\ref{relatedWork}. The system model and our setup are introduced in Section~\ref{preliminaries}. The proposed approach along with the NMIP formulation and the derivation of the detection probability is given in Section~\ref{mpc}. The evaluation of the proposed approach is discussed in Section~\ref{simulationExperiments}, and finally Section~\ref{conclusion} concludes the paper and discusses future research directions.

	\section{Related Work}
	\label{relatedWork}
	Castaway search in post shipwreck disasters has been examined in several works in the past, assuming a number of different setups. Xinming Hu et al.~\cite{hu2020research} explored the use of Kalman Filtering (KF)~\cite{welch1995introduction} to predict the castaway position and a Dynamic Window Approach (DWA)~\cite{fox1997dynamic} for the Unmanned Surface Vehicle's (USV) path planning and obstacle avoidance. Others have investigated the use of both UAV and USV for aiding in the rescue and response. Xiao Xuesu et al.~\cite{xiao2017uav} introduced a UAV for the use of an overhead view camera, helping in the state estimation of the USV. Thus, the system is able to autonomously navigate while First Responders (FRs) can focus on task-level needs. Anibal Matos et al.~\cite{matos2016multiple} took a different approach to the multi-robot solution, utilizing a USV for a first large-scale bathymetry of the interest area and then deploying an Autonomous Underwater Vehicle (AUV) for the close inspection of suspicious targeted areas. In our work, we utilize multiple instances of KF for estimating the state of each castaway. We also employ a single UAV for its maneuverability and its ability to rapidly adapt between flying and hovering.
	
	On the other hand, Ramirez et al.~\cite{ramirez2011coordinated} proposed an expert system utilizing a UAV and a USV for maritime SAR operations with the UAV acting as a remote fast moving sensor that assists the USV during the mission by providing measurements about the castaways thus, letting the USV correct their estimated position using a Particle Filter (PF)~\cite{gustafsson2010particle} for each one of them. As mentioned previously, we use KF instead, purely for its lower computational needs since using multiple PF on an embedded device is computationally intensive.
	
	The use of Non-linear Programming (NLP) has also been used widely over the years. Some have investigated an Artificial Potential Field (APF) path planning technique for producing flyable paths for multi-rotor UAVs in ground target tracking systems~\cite{9234396} and obstacle avoidance systems~\cite{9790230}. Such techniques are beneficial for ground moving targets but, their use has mainly been researched for tracking smooth paths. In addition, Dai and Cochran~\cite{5159914} investigated the generation of cooperative paths for UAVs that are parameterized under the Cornu-Sprials. Such paths however, are mostly suitable for fixed-wing aircrafts. Non-linear and Linear MPC have also been exhaustively researched in terms of target tracking, but most of the work has mainly been focused on ground vehicles~\cite{8786418}.  
	
	Many efforts have also been done for solving the problem of UAV control in 3D environments. Yang and Sukkarieh~\cite{4650637} exploited rapidly-exploring random trees (RRTs) for collision avoidance in cluttered environments. Tisdale et al.~\cite{5069834} proposed a decentralized online path planning framework for a team of fixed wing UAVs with the purpose of search and locate missions. In contrast, our work focuses on optimal control of a single UAV in 3D environment for multiple target tracking. By controlling the UAV in all 3 dimensions, gives the ability to improve the performance of a target tracking system, by unlocking the capability to choose between the quality of the observation and the effective sensing range.
	
	In addition, research has also been done for calculating the Probability of Success (POS) of such search missions. Donatien et al.~\cite{agbissoh2019decision} explained in their work how using optimal search theory and analyzing the Probability of Containment (POC) and Probability of Detection (POD) at the area of interest, one can calculate the POS of the operation. 
	
	Finally, closely related to our work is the approach presented in~\cite{liu2017model}. Chang Liu and Karl Hedrick, also proposed a framework that has the objective of minimizing the covariance of the target's estimated location, by using an MPC to control the agent's movements. However, their work was limited to a two-dimensional environment and to a single ground target. Moreover, the authors did not consider any detection probability assumptions in the sensing model of the agent.
	
	In this paper, we propose an MPC formulation for multiple target tracking missions utilizing a UAV agent for tracking multiple castaway targets that are assumed to be drifting afloat in the aftermath of a maritime accident. To achieve this, in the proposed approach we take into account the target detection probability which is learned from real-world data, as well as the inherent uncertainty in the target dynamical and measurement model. Leveraging multiple Kalman Filter (KF) to estimate each castaways' drift at sea, the proposed system is able to predict the castaways' location and provide informed position inputs to the UAV agent's control system. Then, by moving towards those targets, the system minimizes the measurement noise and simultaneously aids the throughput of Computer Vision (CV) techniques that are usually used in such missions. As a note, and to better simulate a real-life CV detector, we have experimentally analyzed the target detection probability using two different sized and colored water buoys, by training various CNNs. From the evaluation of the CNNs, we constructed a piece-wise linear function that describes the confidence of detection probability based on the UAV's altitude. Overall, the system aims at tracking all castaways, giving the FRs actionable information about the survivors. The proposed technique considers the UAV's kinematic and sensing constraints, and optimizes the agent's control actions over a rolling finite planning horizon, which aim at minimizing the overall tracking uncertainty. 	
	 
	\section{Preliminaries}
	\label{preliminaries}

	\subsection{System Architecture}
	\label{systemInteg}
\begin{figure}
		\centering
		\includegraphics[width=\columnwidth, height=120pt]{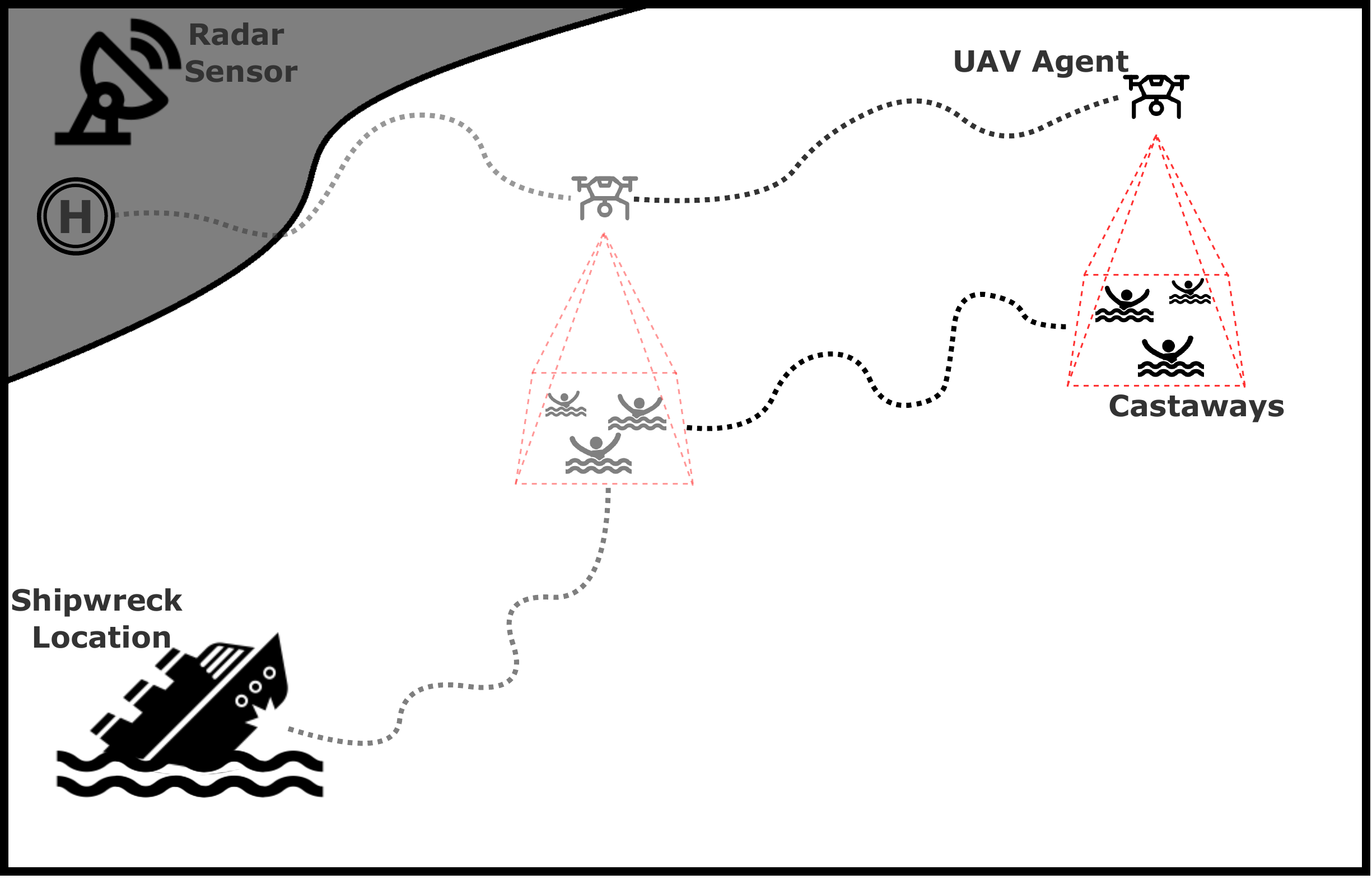}
		\caption{Multiple castaway track scenario illustration at two time steps}
		\label{fig:scenarioPlot}
		\vspace{-12pt}
	\end{figure}

	For the proposed architecture we consider two main components, the mission initiating system (i.e. a coastal radar) and a UAV agent taking over the mission after initially being alerted by the former component. Starting from the initial sensor measurements, we initialize one Kalman Filter (KF) \cite{welch1995introduction} for every target with an initial prior on their states received from the coastal radar. As long as the castaways are within the field of view (FoV) of the onboard camera sensor (further defined in Sec.~\ref{agentDynamics}), the KF uses the measurement from the onboard camera sensor to correct the prediction. When a castaway is not within the FoV of the UAV's onboard camera, then the KF is used to update the location estimates but without the correction steps. While the KF correction steps are not used, the uncertainty of their location increases. In order to minimize the uncertainty and get a better estimate of the castaways' locations, the UAV has to observe the targets with its camera's limited sensing range. However, detecting the targets is stochastic and depends on the probability of detection (described in Sec.~\ref{probDetection}), which is inversely proportional to the agent's altitude. An illustration of the described scenario can be seen in Fig.~\ref{fig:scenarioPlot}.

	\subsection{Agent Dynamics and Sensing Model}
	\label{agentDynamics}
	In our work, the UAV agent is free to move in the 3D Cartesian space and its movement is governed by the following discrete time, linear dynamical model:	
	\begin{equation}
		\label{eq:agentTransition}
		\chi^a_{k+1}=A \chi^a_{k}+ Bu^a_{k}
	\end{equation}
	where $\chi^a_k = [p^a_k, v^a_k]^T$ is the state of the agent at the $k^{th}$ time step which is consisted from its position $p^a_k=[x^a_k, y^a_k, z^a_k]^T \in \mathbb{R}^{3\times1}$ in three-dimensional Cartesian coordinates and its velocity $v^a_k=[\dot{x}^a_k, \dot{y}^a_k, \dot{z}^a_k]^T \in \mathbb{R}^{3\times1}$. The term $u^a_{k} =  [u^a_x, u^a_y, u^a_z]^T \in \mathbb{R}^{3\times1}$ denotes the input control force for each dimension while the system matrices $A \in \mathbb{R}^{6\times6}, B \in \mathbb{R}^{6\times3} $ are given below:
	\begin{gather}
		A=
		\begin{bmatrix}
			\mathbf{I}_{3\times3} & \delta t \mathbf{I}_{3\times3}\\
			\mathbf{0}_{3\times3} & \rho \mathbf{I}_{3\times3}\\
		\end{bmatrix}
		,B=
		\begin{bmatrix}
			\mathbf{0}_{3\times3} \\
			\xi \mathbf{I}_{3\times3}
		\end{bmatrix}
	\end{gather}
	The term $\delta t$ denotes the sampling interval while, $\mathbf{I}_{3\times3}$ and $\mathbf{0}_{3\times3}$ denote the identity and zero matrix, respectively. The parameter $\rho \in [0,1]$ is used to model air resistance whereas the parameter $\xi = {\frac{\delta t}{m}}$ is used to convert the input control force to acceleration, where $m$ is the agent's mass.
	
	As mentioned previously, the agent is carrying an onboard camera sensor. The onboard camera is capable of detecting the castaways and measuring their 2D location i.e., Cartesian (x,y)-coordinates. The effective sensing area of the onboard camera is defined by a rectangle projected on the sea surface which's size is further defined by the horizontal ($\theta_h$) and vertical ($\theta_v$) FoVs in degrees. The horizontal ($l_h$) and vertical ($l_v$) lengths of the rectangle are calculated by $l_{h \mid k} = 2 z^a_k \tan\left(\frac{\theta_h}{2}\right)$ and $l_{v \mid k} = 2 z^a_k \tan\left(\frac{\theta_v}{2}\right)$.
	
	 The measurement provided by the sensor is defined by the random set $Z_k$, which can be either empty with probability $1-p_k, p_k \in \left[0,1\right]$  or be a singleton set with probability $p_k$ (see Sec.~\ref{probDetection}) with its element distributed over the state space of the target $\chi^{c_i}_{k}$, where $c_i$ refers to the $i^{th}$ castaway, according to the probability density function (PDF) $pr(\chi^{c_i}_{k})$. Therefore, the PDF of $Z_k$ is given by:
	\vspace{-3pt}
	\begin{flalign}
		&f\left(Z_k\right) = \left\{ 
		\begin{matrix}
		1-p_k & \hspace{-7pt}\text{if } Z_k=\emptyset&\\
		\vspace{2pt}
		p_k pr(\chi^{c_i}_{k}) & \quad\text{if } Z_k=\left\{y_{k}^{c_i}\right\}
		\end{matrix}\right.&
	\end{flalign}
	 
	The single target measurement likelihood function is further given by $y_{k}^{c_i} = h(\chi^{c_i}_{k}) + w(p_{k})$ where:
	\begin{flalign}
		&h(\chi^{c_i}_{k}) = 
		\begin{bmatrix}
				1 & 0 & 0 \\ 0 & 1 & 0
		\end{bmatrix}\chi^{c_i}_{k}, &w(p_{k}) \sim  \mathcal{N}(0, p_k^{-1} \gamma)&
	\end{flalign}
	$w(p_{k})$ is the measurement noise, which is randomly distributed by a zero mean Normal density with standard deviation $p_k^{-1} \gamma$, where $\gamma$ is the scaling parameter.
	
	\subsection{Target detection probability}
	\label{probDetection}
\begin{figure}
		\includegraphics[width=\columnwidth]{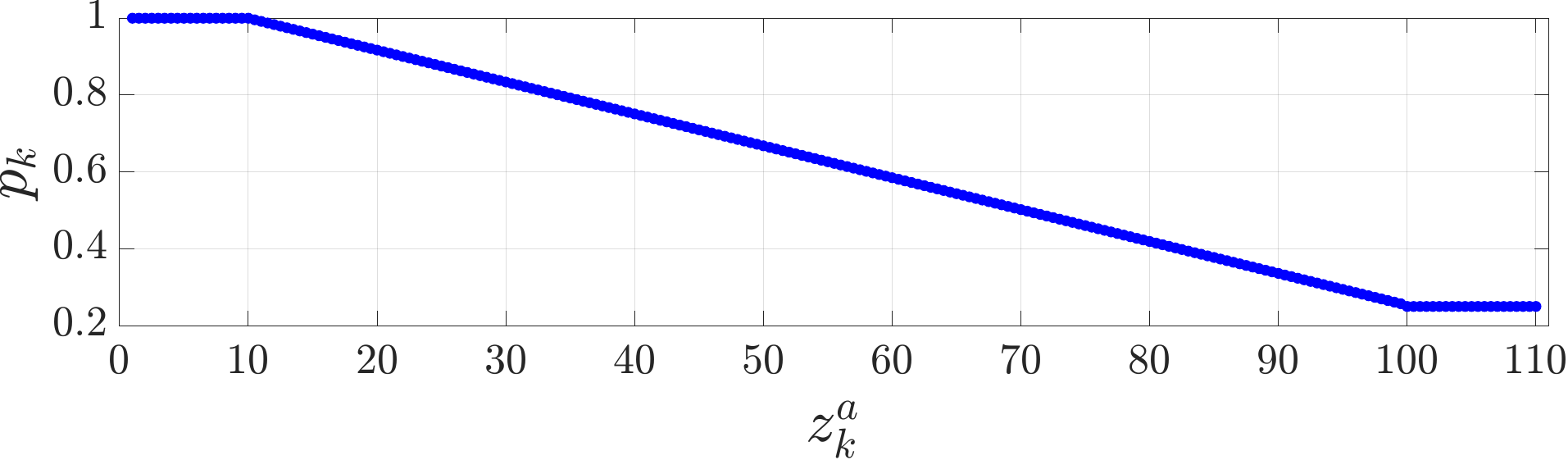}
		\caption{Piece wise function emulating the probability of receiving a measurement from the sensor. Parameters were set to  $\alpha_1=10, \alpha_2=100$ and $\beta_1=-0.0083, \beta_2=1.083$}
		\label{fig:pwf}
		\vspace{-12pt}
	\end{figure}

	This subsection elaborates on the derivation of the probability of detection which we assume to follow a piece-wise linear function. As previously mentioned in Sec.~\ref{agentDynamics}, we receive a non-empty measurement from the onboard camera sensor with probability $p_k$. Thus, probability $p_k$ is assumed to obey the following piece-wise function(see Fig.~\ref{fig:pwf}):
	\begin{flalign}
		&p_{k} = \left\{
		\begin{matrix}
		\vspace{2pt}
		&1 &\text{if } z^a_k\le\alpha_1\\
		\vspace{2pt}
		&p_{min} &\text{if } z^a_k\ge\alpha_2\\
		&\beta_1 z^a_k +\beta_2 &\text{otherwise}
		\end{matrix}\right.&
	\end{flalign}
	where, parameters $\alpha_1$ and $\alpha_2$ are used to tune the altitudes where the probability gets at its maximum and its minimum value. Parameters $\beta_1$ and $\beta_2$ are used to define the slope of the linear function. Note that these tuning parameters must be chosen as such as to not violate the probability rule of $p_k \in [0,1]$. The above modeling assumptions have been verified experimentally, as discussed next in Sec.~\ref{mpc}.	

	\subsection{Castaway Motion Model}
	\label{castMotion}
\begin{figure}
		\centering
		\includegraphics[width=0.75\columnwidth]{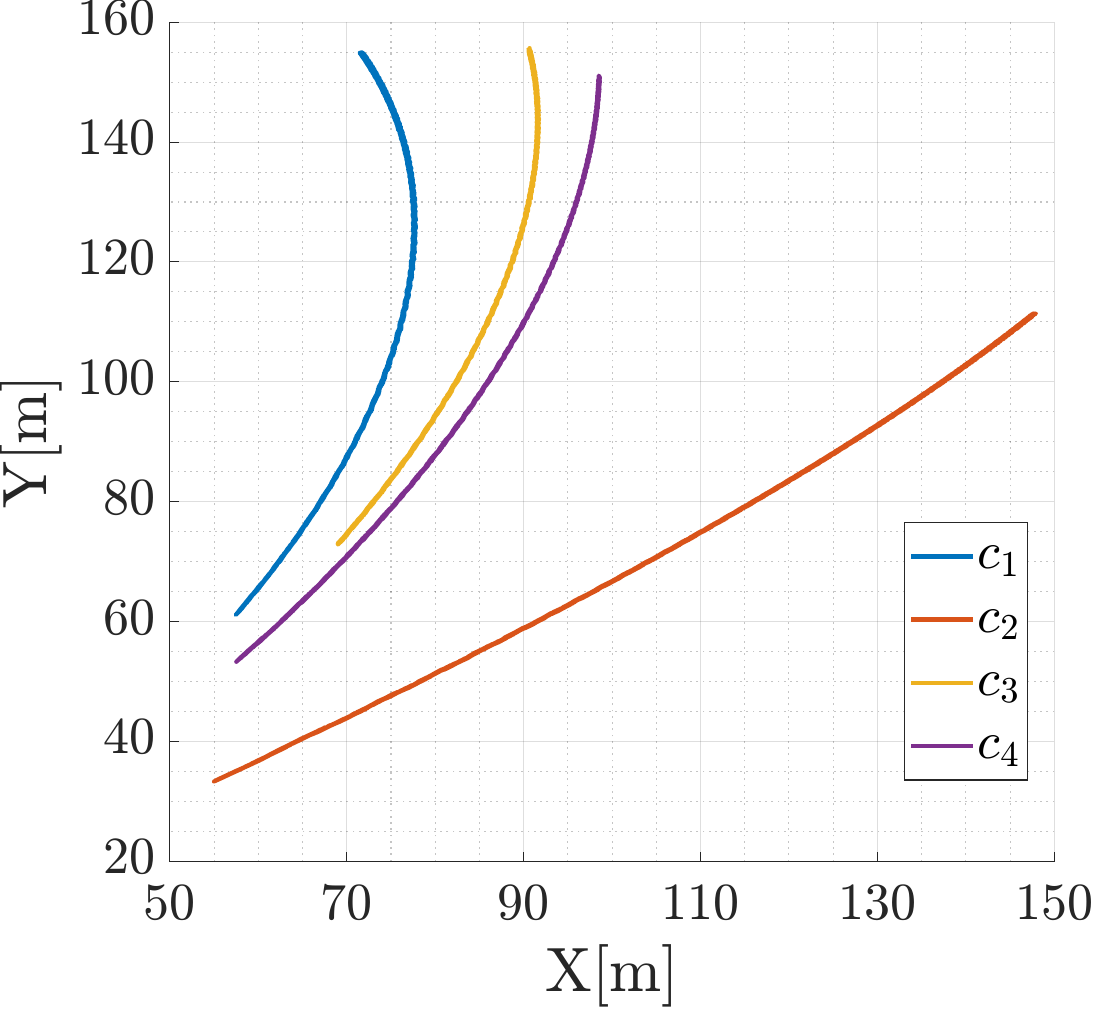}
		\caption{Drift paths of four castaways induced by small amplitude waves for the span of an hour.}
		\label{fig:scen1Plot}
		\vspace{-12pt}
	\end{figure}

	Each castaway is moving in 3D space and its motion is governed by a modified version of the Stokes' drift equations~\cite{shen2001theoretical}. The equations are used to model and generate the ground truth states of the castaways $\chi^{c_i}_{k} = [x^{c_i}_k, y^{c_i}_k, z^{c_i}_k]^T, \forall i=[1,..,\mathcal{C}]$ given by~\eqref{eq:castaway}. In partiqular Eq.~\eqref{eq:waveq} describes the water velocity in the direction of the wave propagation that a specific castaway experiences. Thus, the castaway's location is updated as show in eq.~\eqref{eq:castStateUpdate}.
	\begin{subequations}
		\label{eq:castaway}
		\begin{flalign}
			&v^{c_i}_{k} =  {\frac{\omega H}{2}} e^{wd^{c_i}} \sin{(qd^{c_i}-\omega k)}  \label{eq:waveq}&\\
			&\chi^{c_i}_{k+1} = \chi^{c_i}_{k} + {v^{c_i}_{k}} \begin{bmatrix}
			\cos(\varphi) & \sin(\varphi) & 1
			\end{bmatrix}^T dt&\label{eq:castStateUpdate}
		\end{flalign}
	\end{subequations}
	where $\omega={2\pi}/T$ is the wave frequency, $H$ is the wave height, $w$ is the wave decay-rate and $d^{c_i}$ is the distance of the $i^{th}$ castaway from the wave source. The wave number is given by $q=2\pi/L$ and depends on the wavelength $L$. The angle between the castaway and the wave source is given by $\varphi$ while the wave period is given by $T=\sqrt{2\pi L / gK}$ where, $K=\tanh{(qD)}$, $D$ is the water depth and $g$ is the gravitational acceleration. The above wave equation in~\eqref{eq:waveq} assumes small amplitude waves ($q H \ll 1$), deepwater conditions ($K<1$) and no wave reflections. The ground truth drift of multiple castaways is shown in Fig.~\ref{fig:scen1Plot}. The figure shows the drift of 5 castaways induced over the period of 60 minutes.	

	\section{Proposed Approach}
	\begin{figure}
		\includegraphics[width=\columnwidth, height=120pt]{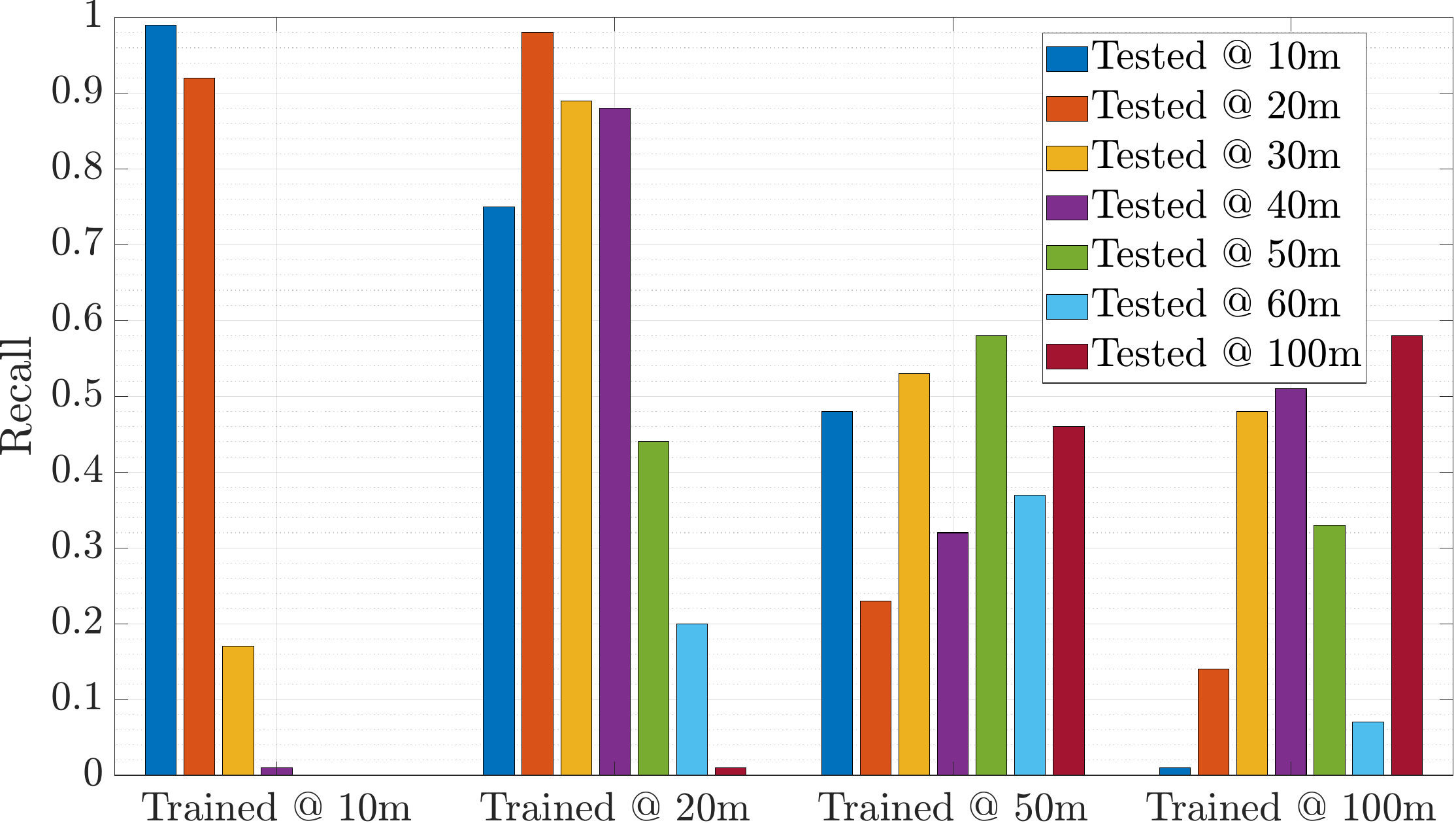}
		\caption{Recall of each CNN at every altitude.}
		\label{fig:cnn}
		\vspace{-10pt}
	\end{figure}
	
	\begin{figure*}
		\includegraphics[width=\textwidth, height=120pt]{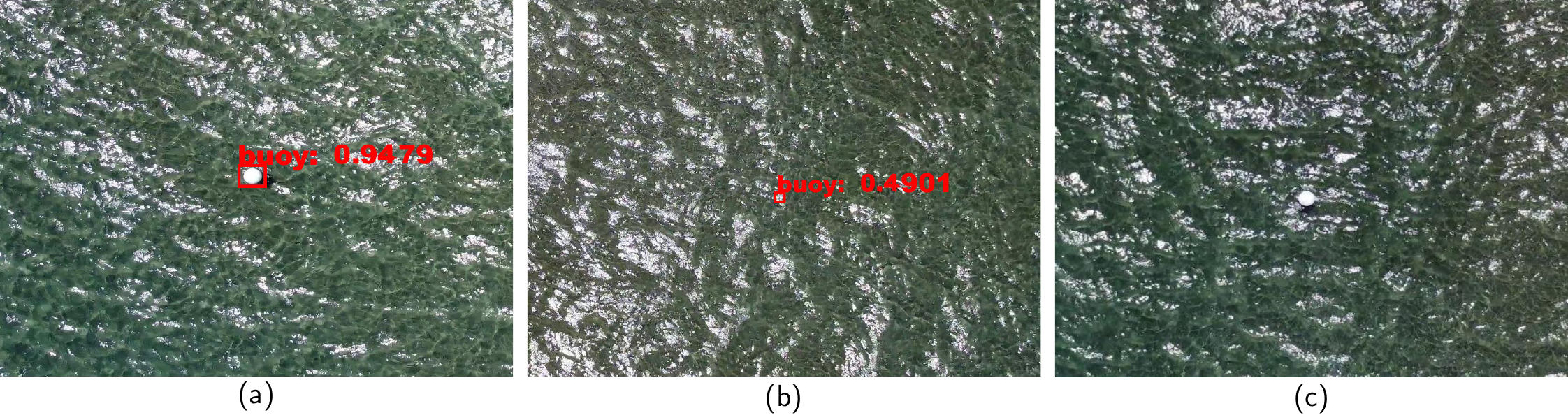}
		\caption{Sample images of detected buoys at 30 meters altitude. Image (a) shows a successfully detected buoy. Image (b) shows a false positive detection in the absence of buoys. Image (c) shows a false negative detection in the presence of a buoy.}
		\label{fig:buoy}
		\vspace{-13pt}
	\end{figure*}	
	
	\subsection{Determining the Target Detection Probability}
	To better determine a realistic target detection probability, the proposed piece-wise linear function was validated by real-life data of water objects and more specifically buoys. We collected more than 10 thousand images of two different types of buoys (white and red) in the city of Larnaca in the Republic of Cyprus. The images were taken using an off-the-shelf UAV carrying a 4K camera from the altitudes of 10-100 meters in two different sea conditions. The first set of images was taken during a calm sea with zero wind while the second set, was taken during small waves and windy conditions. Each image was manually labeled for training a single class detector. Four Convolutional Neural Networks (CNNs) were trained at detecting buoys from the various altitudes. The initial resolution of the images was $1920\times1080$ pixels and the custom dataset was split into four groups:
	\begin{itemize}
		\item[$1^{st}$:] Trained with 2300 images taken from 10 meters altitude and validated with roughly 300 images and tested with roughly 300 images.
		\item[$2^{nd}$:] Trained with 2900 images taken from 20 and 30 meters altitude and validated with roughly 370 images and tested with roughly 370 images.
		\item[$3^{rd}$:] Trained with 3700 images taken from 50 and 60 meters altitude and validated with roughly 470 images and tested with roughly 470 images.
		\item[$4^{th}$:] Trained with 1400 images taken from 100 meters altitude and validated with roughly 200 images and tested with roughly 200 images. 
	\end{itemize}
	Manually labeling the images resulted to more than 14 thousand labeled buoys. The images in the custom dataset were split into 80\% for training, 10\% for validation and 10\% for testing. For training, we utilized the AlexeyAB's version of the Darknet framework~\cite{2004.10934} by using the YOLOv4-tiny CNN mainly for its real-time capabilities. During training, all the default network parameters were chosen and the YOLOv4-tiny's default augmentation configurations were used. The training image size was set to 416x416 while 200 epochs of batch 32 for each network were done. Another functionality given by the YOLOv4 framework called random resizing was also used. Random resizing as its name suggests, randomly resizes the network input size every 10 batches (iterations) form scale 0.7 to scale 1.4 while keeping the initial aspect ratio of the image. Training of the networks was done on an NVIDIA Tesla v100 GPU, with the validation set being used during training. After training, each network was tested with all the test sets, at all altitudes. This was done to derive each networks ability of detection when given images that contain at least one buoy. During testing, the number of true and false positives as well as the amount of false negatives for each network at every altitude were counted. Fig.~\ref{fig:cnn} shows the recall that each network achieved at every altitude. Recall is calculated by dividing all true positives over all the true positives and false negatives. It can be seen that our previous assumption of decreasing confidence with increasing altitude was logical and close to truth. Thus, from Fig.~\ref{fig:cnn} we constructed the confidence of detection function in eq.~\eqref{eq:DetectConf} which we use in the proposed solution to mimic the behavior of a traditional CNN at detecting overboard castaways. A sample of three images of detected buoys can be seen in Fig.~\ref{fig:buoy}. The images were taken at an altitude of 30 meters. In the first image(a), a true positive detection of a white buoy can be seen during small amplitude waves. In the second image (b), a single false positive detection can be seen in the absence of buoys. Finally, in the last image (c) a false negative detection can be seen where the detector failed to recognize the white buoy.	
	\begin{flalign}
		\label{eq:DetectConf}
		&f^d(z^a_k) = \left\{
		\begin{matrix}
			\vspace{2pt}
			&1 &\text{if }z^a_k\le10\\
			\vspace{2pt}
			&0.25 &\text{if }z^a_k\ge100\\
			&-0.0083 z^a_k +1.083 &\text{otherwise}
		\end{matrix}\right.&
	\end{flalign}

	\subsection{Model Predictive Control Formulation}
	\label{mpc}
	To compute the vector $u^a_{k}$ for controlling the UAV's movement, a receding horizon MPC is used. The MPC is formulated such that it does not infringe the UAV's capabilities and limitations such as horizontal and vertical speeds and accelerations. To achieve that, we formulate an optimization problem that minimizes the trace of the state covariance matrix $P^{c_i}_{k+\tau \mid k}$ for each castaway's predicted location by controlling the input of the agent as referred at Sec.~\ref{agentDynamics}. In this work, the KF equations are employed to be able to predict the castaway's drift and provide estimates of its location and the accumulated uncertainty of the open-loop system (when no measurements are received over an extended period of time) over the planned horizon. After being initialized with noisy measurements about the castaway's state, the KF equations predict the castaway's next state and then correct the estimate using pseudo-measurements created by adding noise to the predicted positions, whenever the castaway is assumed to be within the FoV. The proposed model predictive controller, is shown in Eq, \eqref{eq:mpcObj}-\eqref{eq:sets}, as a non-linear mixed integer program.
	
	The objective in eq.~\eqref{eq:mpcObj} is to minimize the trace of the covariance matrix $P^{c_i}_{k+\tau \mid k}$ of each castaway over all the time steps of the planning horizon where, $N$ is the number of steps in the receding horizon and $\mathcal{C}$ is the number of castaways. Eq.~\eqref{eq:agentMotion} defines the UAV's dynamical model as previously discussed in Sec.~\ref{agentDynamics}. Vector $u^a_{k+\tau \mid k}$ is the control input of the UAV consisting of $(u^a_x,u^a_y,u^a_z)$ for all steps in the horizon $\tau\in \left\{0,...,N-1\right\}$. Constrain~\eqref{eq:fovSize} refers to the derivation of the horizontal and vertical size of the onboard camera's FoV.
	\begin{subequations}\label{eq:mpc}
		\begin{equation}
			\label{eq:mpcObj}
			\arg\min_{\hspace{-19pt}u^a_{k+\tau \mid k}} \sum_{i=1}^{\mathcal{C}}\sum_{\tau=0}^{N-1}tr(P^{c_i}_{k + \tau \mid k})\\
		\end{equation}
		\begin{flalign}
			&\text{subject to } i \in\left\{1,...\mathcal{C}\right\}, \tau\in \left\{0,...,N-1\right\} \text{:} \notag&
		\end{flalign}
		\begin{flalign}
			\label{eq:agentMotion}
			&\chi^a_{k+\tau+1 \mid k}=A^a \chi^a_{k+\tau \mid k}+ Bu^a_{k+\tau \mid k}& \forall \tau\\
			\label{eq:fovSize}
			&l^{j}_{k+\tau \mid k} = z^a_{k+\tau \mid k} \tan({\theta_j})& \forall j, \tau\\
			\label{eq:fovLimits}
			&\mathrm{d}   ^{j}_{k+\tau\mid k} = D_j \chi^a_{k+\tau\mid k} + (-1)^j l^{j}_{k+\tau\mid k}&\forall j, \tau\\
			\label{eq:binariesFOV}
			&b^{{c_i} \mid j}_{k+\tau \mid k}=\left\{\begin{matrix}
				1, &\text { if } E_j \chi^{c_i}_{k+\tau \mid k}\leq \mathrm{d}^{j}_{k+\tau \mid k} \\
				0, &\text { otherwise }\\
			\end{matrix}\right.&\forall i,j, \tau\\
			\label{eq:binSum}
			&\mathrm{s}^{c_i}_{k+\tau \mid k}=\sum_{j=1}^4 b^{{c_i}\mid j}_{k+\tau \mid k}&\forall i, \tau\\
			\label{eq:binFoV}
			&\mathrm{b}^{c_i}_{k+\tau \mid k}=\left\{\begin{matrix}
				1, &\text { if } \mathrm{s}^{c_i}_{k+\tau  \mid k} = 4 \\
				0, &\text { otherwise }\\
			\end{matrix}\right.&\forall i, \tau\\
			\label{eq:priPos}
			&\chi_{k+\tau+1 \mid k}^{c_i}=A^c \hat{\chi}_{k+\tau \mid k}^{c_i}&\forall i, \tau\\
			\label{eq:priCov}
			&P_{k+\tau+1 \mid k}^{c_i}=A^c \hat{P}_{k+\tau \mid k}^{c_i} {A^c}^{T}+Q&\forall i, \tau\\
			\label{eq:gain}
			\begin{split}
				K_{k+\tau+1 \mid k}^{c_i}&=\hat{P}_{k+\tau+1 \mid k}^{c_i} C^{T}\\
				&\left(C \hat{P}_{k+\tau+1 \mid k}^{c_i} C^{T}+R_{k+\tau \mid k}\right)^{-1}\\
			\end{split}&\forall i, \tau\\
			\label{eq:postPos}
			\begin{split}
				\hat{\chi}_{k+\tau+1 \mid k}^{c_i}&=\chi_{k+\tau+1 \mid k}^{c_i}+\mathrm{b}_{k+\tau \mid k}^{c_i} K_{k+\tau+1 \mid k}^{c_i}\\
				&\left(y_{k+\tau+1 \mid k}^{c_i}-C \hat{\chi}_{k+\tau+1 \mid k}^{c_i}\right)\\
			\end{split}&\forall i, \tau\\
			\label{eq:postCov}
			\begin{split}
				\hat{P}_{k+\tau+1 \mid k}^{c_i}&=P_{k+\tau+1 \mid k}^{c_i}-\mathrm{b}_{k+\tau \mid k}^{c_i} K_{k+\tau+1 \mid k}^{c_i}\\
				&C P_{k+\tau+1 \mid k}^{c_i}\\
			\end{split}&\forall i, \tau\\
			\label{eq:measureNoise}
			&\sigma_{k+\tau \mid k} = \gamma r(z^{a}_{k+\tau \mid k})&\forall \tau\\
			\label{eq:measureCov}
			&R_{k+\tau\mid k}=\mathbf{I}_{2\times2}\sigma_{k+\tau \mid k}&\forall \tau\\
			\label{eq:psudomeasure}
			&y^{c_i}_{k+\tau+1 \mid k} = C\hat{\chi}^{c_i}_{k+\tau+1 \mid k} + n(\sigma_{k+\tau \mid k})&\forall i, \tau\\
			\label{eq:forceSmoothing}
			&\delta u^a_{k+\tau+1 \mid k} = (u^a_{k+\tau+1 \mid k} - u^a_{k+\tau \mid k})^2&\forall \tau\\
			\label{eq:boundaries}		
			&\chi^a_{k \mid k} \in \mathcal{X}, u^a_{k \mid k} \in \mathcal{U}, \delta u^a_{k \mid k} \in \delta \mathcal{U}&\\
			\label{eq:binaryLims}
			&b^{{c_i} \mid j}_{k+\tau \mid k}, \mathrm{b}_{k+\tau \mid k}^{c_i} \in \left\{0,1\right\}&\forall j, i, q, \tau\\
			\label{eq:integerLims}
			&\mathrm{s}^{c_i}_{k+\tau \mid k} \in [0,..,4]&\forall i,\tau\\
			\label{eq:sets}
			&j=[1,..,4], q=[1,2,3]&
		\end{flalign}
	\end{subequations}
	 For simplicity, the horizontal and vertical FoV have been split to halves. Thus, constrain~\eqref{eq:fovLimits} can now be used to estimate the left, right, top and bottom limits of the FoV in the 2D Cartesian plane. Matrix $D_j \in \mathbb{R}^{1\times6}$ is used to select the X or Y location from the agent's state vector. Using the FoV limits from~\eqref{eq:fovLimits}, we use constrains~\eqref{eq:binariesFOV},~\eqref{eq:binSum} and~\eqref{eq:binFoV} to calculate whether a target is within the FoV or not. Do do that, we check if a target is to the right of the left limit, to the left of the right limit, below the top limit and above the bottom limit of the FoV. Each check activates a binary variable~\eqref{eq:binariesFOV} using the big M technique, and the inclusion within the field of view is done by the summation of all these variables~\eqref{eq:binSum}. Matrix $E_j \in \mathbb{R}^{1\times3}$ is used to select the X or Y location from the target's state vector. If the summation of the binaries is equal to 4~\eqref{eq:binFoV}, then another binary variable is activated, indicating that the target is within the FoV. In addition, constrains~\eqref{eq:priPos} to~\eqref{eq:postCov} describe the KF estimator. Constrain~\eqref{eq:priPos} regards to the target's a priori state estimate using the transition matrix $A^c=
	\begin{bmatrix}
		\mathbf{I}_{2\times2} & \delta t \mathbf{I}_{2\times2} \\ \mathbf{0}_{2\times2} & \mathbf{I}_{2\times2}
	\end{bmatrix}$.
	The matrices $\mathbf{I}_{2\times2}$ and $\mathbf{0}_{2\times2}$ represent the identity and zero matrix respectively while $\delta t$ defines the sampling interval. We use the KF only for the (X,Y) location of the target in the 2D Cartesian plane to reduce the complexity of the system. Estimating the target's position in the Z axis can be neglected due to its very small variations. Constrain~\eqref{eq:priCov} computes the a priori covariance matrix. Matrix $Q$ is constant and it represents the covariance of the process noise. Constrain~\eqref{eq:gain} computes the KF gain. Due to the inverse in this constrain, we needed to use non linear constrains in our formulation. Subsequently, constrain~\eqref{eq:postPos} calculates the posteriori state estimate of the target using a psuedo-measurement derived in contrain~\eqref{eq:psudomeasure} and constrain~\eqref{eq:postCov} calculates the posteriori covariance matrix of the target's state estimate. Similar to~\cite{liu2017model} we handle the intermittent observations of each target with the use of the technique described in~\cite{sinopoli2004kalman}. However, we apply the technique within the proposed NMIP formulation with the planned horizon's observation binary variables $\mathrm{b}^{c_i}_{k+\tau \mid k}$. Further, constrain~\eqref{eq:measureNoise} is used to create the measurement's standard deviation using the tuning parameter $\gamma$ and function $r(z^a_{k+\tau \mid k})$ which returns a number in the range $[0,1]$ depending on the agent's altitude simulating the piece-wise function defined in Sec.~\ref{probDetection}. Thus, the measurement's standard deviation increases with the increase of altitude. The covariance of the observation noise is calculated using the aforementioned standard deviation as shown in constrain~\eqref{eq:measureCov}. As explained above, in the planned horizon, we use a pseudo-measurement to estimate the target's location and covariance matrix in the future. The pseudo-measurement is derived in constrain~\eqref{eq:psudomeasure} and it is created by adding $n(\sigma_{k+\tau \mid k})  \sim \mathcal{N}(0,\sigma_{k+\tau \mid k})$, a zero mean Gaussian noise to the a priori state estimate of the castasway's location. Matrix $C \in \mathbb{R}^{2\times3}$ is the observation matrix and is used to get the X and Y position from the target's state vector. Constrain~\eqref{eq:forceSmoothing} is employed to smooth the transition between the forces in the horizon by bounding the squared difference between each two consecutive forces in the horizon. Finally, eq.~\eqref{eq:boundaries} ensures that the UAV is kept within the 3D space boundaries set by the $\mathcal{X}$ and the input vector is within the realistic capabilities $\mathcal{U}$ of the agent.
	
	\section{Simulation Experiments}
	\label{simulationExperiments}
	
	This section elaborates on the simulation setup and procedure which was used to evaluate the proposed approach. Sec.~\ref{simulationSetup} describes the setup used including the UAV's model parameters and the computation of the castaways ground truth trajectories. Using this setup,  Sec.~\ref{simulationResults} provides simulation results and elaborates on key findings and insights.
	
	\subsection{Simulation Setup}
	\label{simulationSetup}
	As described in section~\ref{castMotion}, we create ground truth castaway drifts that are subsequently used to evaluate our framework in simulation. The scenario considered the drift induced after one hour of simulation, with randomly generated wave sources of various amplitudes $H$, decay-rates $c$ and wavelengths $L$. For all runs the water depth $D$ was constant and both small amplitude ($q\cdot H \ll 1$) and deepwater ($K<1$) conditions were satisfied. The ground truth drifts used in the simulation are plotted in Fig.~\ref{fig:scen1Plot} where each colored line corresponds to a castaway's path.

	The measurements given by the stationary radar at the beginning of the mission and the measurements given by the onboard camera are generated by adding a zero mean Gaussian noise to the ground truth of each castaway. It should be noted that the variance of the added noise is different between the two sensors and in the case of the onboard camera it is proportional to the agent's altitude, with the radar sensor having the biggest variance. Also, the probability of detection is inversely proportional to the altitude of the UAV agent as described in Sec.~\ref{probDetection}. In this work we do not investigate the problem of data association and assume that the agent is able to distinguish between each target.
	
	For the agent's motion and input boundaries aforementioned in Sec.~\ref{mpc}, we used realistic constrains that were taken from off-the-shelf UAVs readily available in the market. As such, the agent's velocity was limited to $\pm11m/s$ in the horizontal axes and to $\pm3m/s$ in the vertical axis. Acceleration was limited to $\pm2 m/s^2$ for all 3 dimensions while the FoV was set to $69^\circ$ horizontally and $54^\circ$ vertically. Finally, the number of steps in the rolling horizon was set to $N=5$.
	
\begin{figure}
		\includegraphics[width=\columnwidth, height=150pt]{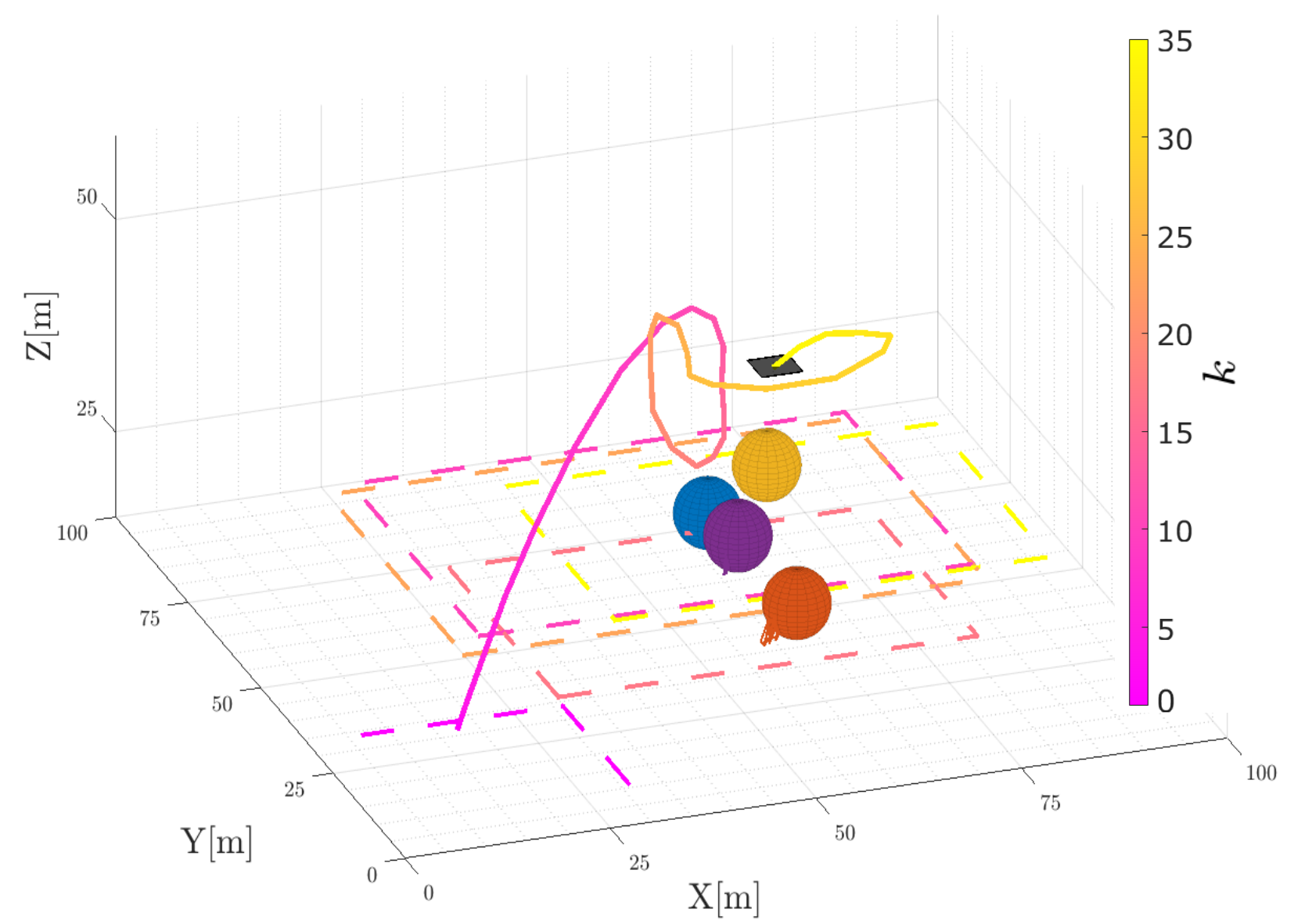}
		\caption{Environment visualization during the first 35 time steps of the operation.}
		\label{fig:simPlot}
		\vspace{-15pt}
	\end{figure}

	A plot of the operation environment is shown in Fig.~\ref{fig:simPlot}. There are four castaways represented by the yellow, blue, purple and orange spheres. The UAV agent is represented with a black square while the onboard camera's sensing area is visualized with the dashed rectangle. In this figure, the sensing area at each one of the five sections is plotted as well as the executed path of the UAV is visualized with the colored line. The colormap represents the change in time.
	
	\subsection{Simulation Results}
	\label{simulationResults}
\begin{figure*}
	\includegraphics[width=\textwidth, height=300pt]{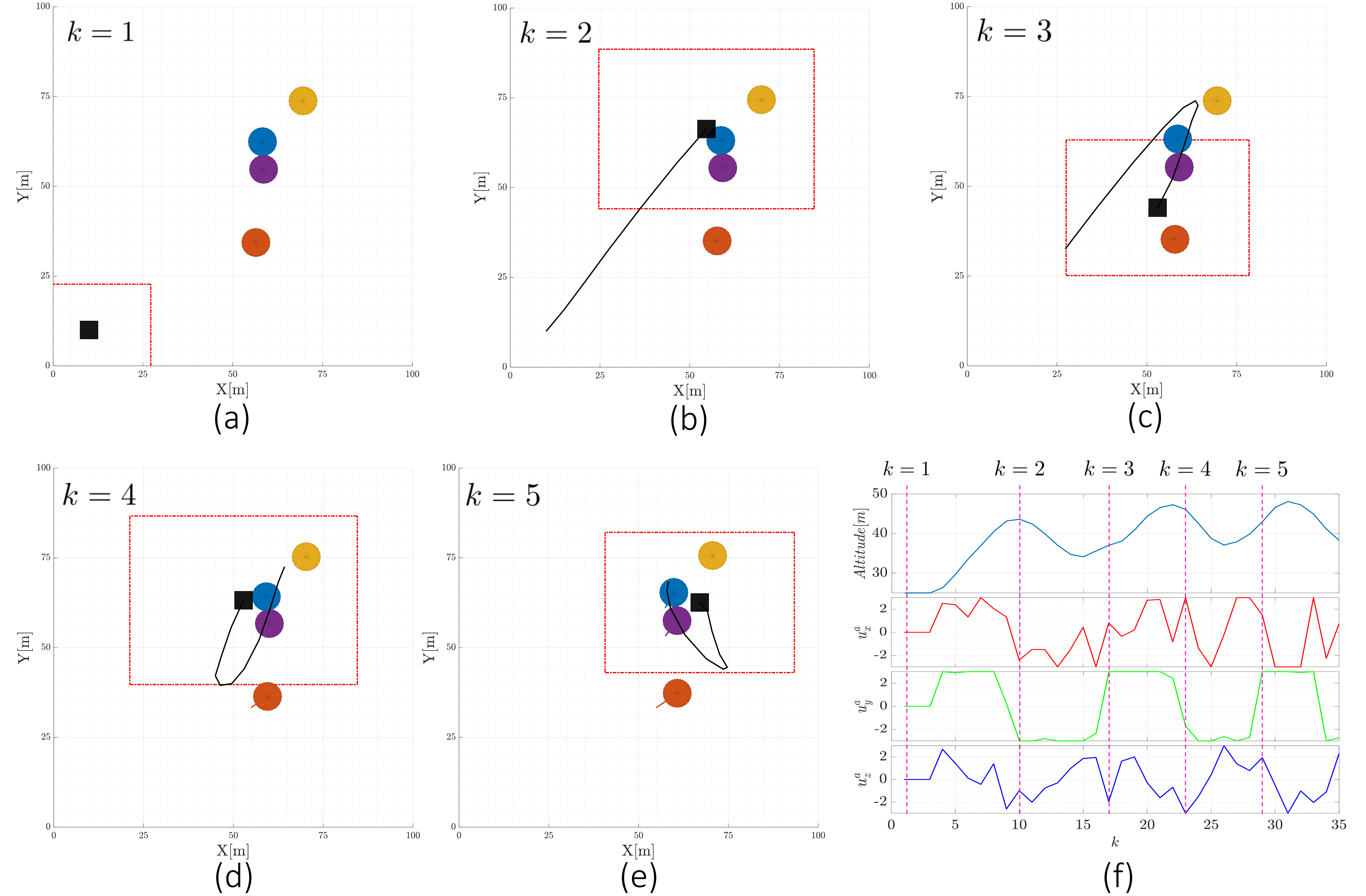}
	\caption{Intermediate steps of the proposed solution. Red rectangle shows the onboard camera's sensing range. Blue, yellow, purple and orange spheres represent the targets. The black line shows the agent's trajectory for the the past 10 time steps.}
	\label{fig:scenPlot}
	\vspace{-15pt}
\end{figure*}

\begin{figure}
		\includegraphics[width=\columnwidth, height=95pt]{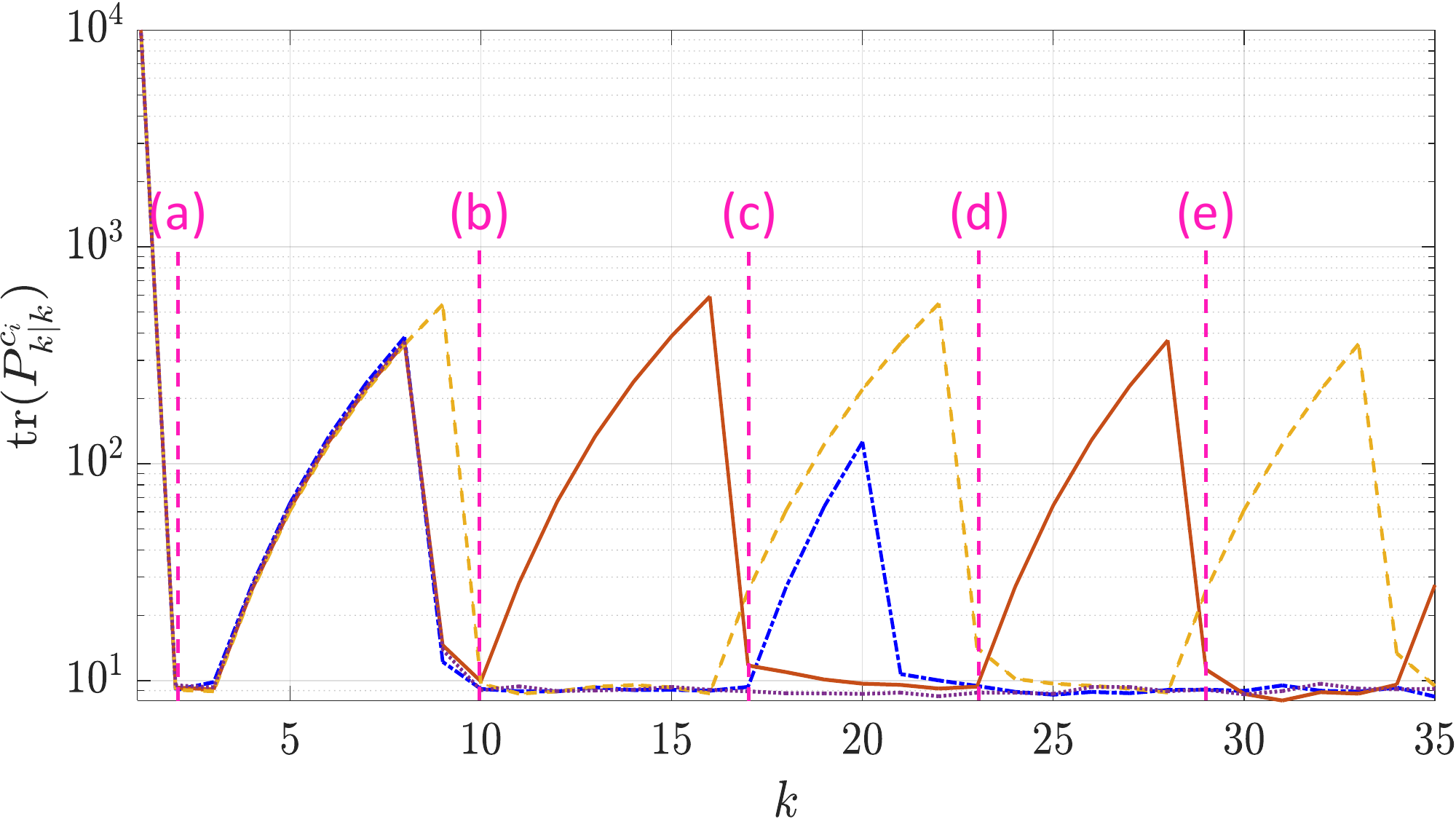}
		\caption{Trace of each target's matrix $P^{c_i}_{k \mid k}$ for the first 35 time steps. Each line represents one of the targets trace covariance value in log scale.}
		\label{fig:covPlot}
		\vspace{-15pt}
	\end{figure}

	Several monte carlo simulations were conducted and in each simulation, we randomly placed the agent's initial position in close proximity to the castaways but not necessarily having the castaway's within the agent's FoV. Figure~\ref{fig:scenPlot} shows the first 35 time steps during the simulation which has been divided into 5 sections (section (a) to (e)). In this figure, the behavior of the agent can be observed. The agent moves back and forth between the four targets with the aim of gathering measurements with the lowest altitude possible. This has the result of minimizing the covariance matrix of each target independently, since the measurement's noise is inversely proportional to the agent's altitude. If the agent cannot traverse between the targets in the given horizon due to physical limitations in its dynamical model, it increases its sensing area by increasing its altitude, thus achieving to cover all the targets. As mentioned in the NMIP objective~\eqref{eq:mpcObj}, our aim is to minimize the trace of the covariance matrix $P^{c_i}_{k+\tau \mid k}$ of each target. As such in the lower right plot of Fig.~\ref{fig:scenPlot} we show both the variation of the agent's altitude along with the three control inputs. It can be easily seen in Fig.~\ref{fig:covPlot}, that the covariance of each target is kept as low as possible during the duration of the simulation. The fluctuation on the value of each trace is caused by the back and forth behavior of the proposed controller. For ease of readability, only the first 35 time steps of the trace value were plotted and the trace values are shown in log scale across the y-axis.
	
	Finally, we have investigated the computational complexity of the proposed approach. In order to do that, we have recorded the average time required to optimally solve the proposed NMIP problem, for various configurations of the planning horizon length ($N$) and number of targets ($\mathcal{C}$), as depicted in Table.~\ref{tbl:execTimes}. Each setup was solved for 1800 times and the cumulative average execution time was recorded. The simulations were conducted on a server machine with an Intel Xeon E5-2680 processor and 256GB of RAM using MATLAB R2020B and Gurobi solver 9.5.2. The results show that the computational complexity increases with the length of the planning horizon and the number of targets. This is attributed mainly to the increase of the number of binary variables that are required by the proposed controller, which increase linearly with the length of the planning horizon and the number of the targets. 
	Although, it is well known that large mixed integer programs (MIPs) can be intractable, recent advances in optimization and machine learning have demonstrated that MIPs of moderate sizes can be solved in real-time~\cite{bertsimas2022online}. A video animation of the whole MATLAB simulation can be accessed at \href{https://youtu.be/B5POJ7GV3-s}{\textit{https://youtu.be/B5POJ7GV3-s}}.
	
\begin{table}[]
	\centering
	\begingroup
	\setlength{\tabcolsep}{10pt} 
	\renewcommand{\arraystretch}{1.25} 
	\caption{Experimental Execution Times [s]}
	\begin{tabular}{c|c|c|}
		\cline{2-3}
		    & $\mathcal{C}=2$      		& $\mathcal{C}=4$      \\  \hline
		\multicolumn{1}{|c|}{$N=3$}		& 0.1290  	& 0.2898   \\  \hline
		\multicolumn{1}{|c|}{$N=5$}		& 0.3897 	& 1.5880   \\  \hline
		\multicolumn{1}{|c|}{$N=7$}		& 1.3814  	& 6.4169   \\  \hline
	\end{tabular}
	\label{tbl:execTimes}
	\endgroup
	\vspace{-10pt}
\end{table}

	\section{Conclusion and Future Work}
	\label{conclusion}
	This work proposed a model predictive control approach for accurately tracking multiple castaways by computing the UAV's control inputs over a rolling planning horizon, which result in the minimization of the estimation error of the states of multiple moving castaway targets. It has been shown, that under the realistic scenario considered in this work, the proposed controller manages to track the multiple castaways reliably under various sea conditions (i.e. wave parameter configurations). Future work will investigate the extension of this approach to multiple UAV agents, and the design of a distributed multi-UAV system for this problem that can be applied in real-world settings.
	
	\section*{Acknowledgment}
	This work is funded by the Cyprus Research and Innovation Foundation under Grant Agreement EXCELLENCE/0421/0586 (GLIMPSE), by the European Union’s Horizon 2020 research and innovation programme under grant agreement No. 739551 (KIOS CoE), and from the Government of the Republic of Cyprus through the Cyprus Deputy Ministry of Research, Innovation and Digital Policy.
	\vspace{-5pt}
{\small
	\bibliographystyle{IEEEtran}
	\bibliography{root.bbl}
}

\end{document}